\documentclass[letterpaper, 10 pt, conference]{ieeeconf}  

\IEEEoverridecommandlockouts                              

\usepackage{amsmath} 
\usepackage{amssymb}  
\usepackage{subfiles}
\usepackage{graphicx} 
\usepackage{balance}
\usepackage{comment}
\usepackage{bm}
\usepackage{adjustbox}
\usepackage{hyperref}
\usepackage[table]{xcolor}

\usepackage{booktabs}

\definecolor{pastelblue}{rgb}{0.68, 0.78, 0.81}
\usepackage{cleveref}

\crefname{section}{Sec.}{Secs.}
\Crefname{section}{Section}{Sections}
\crefname{figure}{Fig.}{Figs.}
\crefname{equation}{Eq.}{Eqs.}
\crefname{table}{Table}{Tables}
\Crefname{figure}{Figure}{Figures}
\crefname{algocf}{Alg.}{Algs.}
\Crefname{algocf}{Algorithm}{Algorithms}

\title{\LARGE \bf

Comparative Analysis of Sampling-Based Trajectory Optimization\\ for 
Deformable Robots
}

\author{Etienne Ménager, Fabian Schramm and Justin Carpentier
\thanks{Authors are with Inria and Département d’Informatique de l’École Normale Supérieure, PSL Research University in Paris, 75013 Paris, France.
France (e-mail: etienne.menager@inria.fr, fabian.schramm@inria.fr, justin.carpentier@inria.fr).}
}

\begin{document}

\maketitle
\thispagestyle{empty}
\pagestyle{empty}

\begin{abstract}
Soft robots exhibit high-dimensional, nonlinear dynamics and frequent contact interactions, making trajectory optimization particularly challenging when costly deformable simulations restrict the available rollout budget. We present a comparative analysis of sampling-based control for these systems without requiring differentiable dynamics. Our study considers a broad family of sampling strategies, from uninformed trajectory sampling to adaptive and locally model-informed approaches. We evaluate them on five tasks spanning stabilization, contact-mediated reaching, object interaction, high-dimensional manipulation, and dynamic motions that exploit delayed mechanical responses and elastic energy storage. Under matched rollout budgets, the analysis highlights that structuring the search is more effective than relying solely on increasingly sophisticated distribution updates. In particular, temporal action parameterizations, coarse-to-fine exploration, preservation of coherent contact modes, and locally guided proposals consistently improve performance in the low-sample regime. The resulting framework, implementation and benchmark provide a reproducible basis for evaluating sampling-based control of complex soft robotic systems.
\end{abstract}


\section{Introduction}
\label{sec:intro}

Soft robots are designed to exploit compliance rather than avoid it. Their bodies can bend, stretch, store energy, make distributed contact with the environment, and generate behaviors that are difficult to achieve with rigid mechanisms~\cite{rus2015design}. This makes soft robots attractive for manipulation, locomotion, and interaction tasks, while also introducing substantial challenges for motion generation. Their dynamics are high-dimensional, strongly nonlinear, and often underactuated. Contact, hysteresis, actuator saturation, and nonsmooth task rewards may further introduce discontinuities or poorly conditioned sensitivities. As a result, controlling these systems while explicitly accounting for their dynamics often requires either sophisticated models or expensive computation.

A natural way to avoid explicit modeling is to learn a controller. Reinforcement learning and other data-driven methods have been used to control soft arms, continuum robots, deformable objects, and soft locomotion systems~\cite{menager:hal-03778189,bhagat2019deep,wang2021mlcontinuum,kim2021reviewmlsoft,zakaria2023drlsoftobject}. These works show that learned policies can account for nonlinearities, hysteresis, and unmodeled effects. However, such policies must be trained, often with many interactions or simulated rollouts, and their performance depends on the exploration strategy, training distribution, and morphology considered. 

Another line of work keeps a model inside the control loop. Recent methods rely on reduced models for control and trajectory optimization, including proxy models, learned compact FEM representations, Koopman embeddings, and physics-informed neural surrogates~\cite{menager2023toward,navez2025modeling,bruder2019koopman,Licher2025AdaptiveMC}. These approaches are powerful because they exploit the structure of the mechanics. However, the controller then depends on the availability of a model that is accurate enough, computationally efficient, and reliable in the region of the state-action space explored by the task. Optimization-based control has also been used to account for contact in soft robotics, but existing formulations are often tailored to one-step control objectives~\cite{coevoet2017optimization,menager2025frictional}.

Gradient-based formulations push this idea further. Differentiable simulators expose derivatives of deformable dynamics for learning, calibration, design, and control~\cite{Hu2019ChainQueenAR,menager2025differentiable}. Derivatives of condensed mechanical models have also been used for trajectory optimization in finite-element simulations of soft robots~\cite{menager2024condensed}. Although powerful, obtaining useful derivatives through deformable dynamics with contact, constraints, and nonsmooth objectives requires dedicated formulations and solvers. Moreover, not all simulation pipelines provide derivatives that are stable, accurate, or available at a reasonable computational cost, which limits their use as the sole source of optimization information. 

This motivates a complementary viewpoint based on shooting and sampling methods. Instead of optimizing over states and controls while enforcing the dynamics as constraints, shooting methods optimize only over action sequences and evaluate candidate trajectories through forward simulation. This makes them compatible with black-box dynamics, easy to parallelize, and robust to nonsmooth rewards or contact events. 
In rigid-body robotics, sampling-based model predictive control (MPC) has recently evolved from a simple derivative-free baseline into a competitive control paradigm, driven largely by the availability of massively parallel simulation. Model Predictive Path Integral control (MPPI) introduced cost-weighted updates of sampled action sequences~\cite{williams2017mppi,Williams2017InformationTheoreticMP}. Predictive Sampling later showed that a much simpler best-sample strategy can remain competitive when combined with warm-starting in a receding-horizon control loop~\cite{Howell2022PredictiveSR}. Recent work has scaled MPPI-like controllers to contact-rich locomotion and whole-body control using full-order dynamics, temporal warm-starts, structured noise, action parameterizations, and parallel rollouts~\cite{xue2024dialmppi,turrisi2024benefits,schramm2025reference}. A recent tutorial gives a unified presentation of these zero-order techniques for robotics~\cite{Armand2025ZeroOrder}. In addition, particular attention has been paid to low-sample regimes, where increasing the number of rollouts is no longer possible. Methods such as covariance adaptation and model-guided proposals have been introduced to concentrate samples in more informative regions~\cite{trevisan2024biased,vlahov2024lowfrequency,schramm2026variancereduced}.

For soft robots, this question remains insufficiently explored. MPPI has started to appear in soft and continuum robotics, but mostly in task-specific settings. MPPI has been applied to tendon-driven continuum robots simulated with Cosserat-rod dynamics and strain parametrization, where the control objective is formulated as a geometric end-effector tracking problem on Lie groups~\cite{arefinia2026errorstate}. Other work applies MPPI to the I$^2$RIS cable-driven surgical snake robot using a data-driven dynamics model corrected online by a Kalman filter to adapt to simulated external disturbances and environmental loads~\cite{esfandiari2025i2ris}. These results show that MPPI is relevant for continuum systems, but they remain tied to a specific robot morphology, model class, and reaching or tracking task. More importantly, they do not investigate whether modern sampling-based controllers can exploit deformation, elastic energy storage, and contact interactions to solve dynamic tasks under the limited rollout budgets typical of soft-robot simulation.

In this paper, we compare sampling-based control for deformable robots under limited rollout budgets, considering both open-loop trajectory optimization and receding-horizon control. Rather than introducing another sampling method, we aim to characterize how existing strategies use the available information and computational budget, and how these choices affect their ability to exploit deformable dynamics. The analysis combines three complementary components:
\begin{itemize}
    \item a benchmark of five deformable robot tasks spanning stabilization, contact-mediated reaching, object interaction, elastic energy exploitation, and higher-dimensional manipulation, evaluated under matched rollout budgets;
    \item a systematic comparison of sampling strategies organized according to the information used to guide the search, from uninformed trajectory sampling to cost-weighted, distribution-adaptive, and locally model-informed approaches;
    \item an analysis of the resulting control behaviors, including contact-mode discovery, delayed mechanical effects, elastic loading and release, temporal action structure, and the effect of replanning and planning horizon.
\end{itemize}

\section{Problem Formulation and Sampling-Based Control}
\label{sec:method}

This section establishes the common formulation used throughout the comparative analysis. We first formulate the finite-horizon control problem and introduce the deformable dynamics used for rollout evaluation. We then present the sampling strategies considered in the benchmark in a common notation, from direct trajectory sampling to distribution adaptation, cost-based reweighting, and locally model-informed proposals. 

\subsection{Finite-Horizon Control Problem}

Let $\bm{x}_t$ denote the system state and $\bm{u}_t$ the control input. For each control task, we assume a reward function 
\begin{equation}
    r_t = r(\bm{x}_t, \bm{u}_t),
\end{equation}
which may include task objectives, control regularization, velocity penalties, contact-dependent terms, or terminal bonuses. Given an initial state $\bm{x}_0$ and an action sequence 
\begin{equation}
\label{eq:action_sequence}
    \bm{U} = (\bm{u}_0, \ldots, \bm{u}_{T-1}),
\end{equation}
a rollout through the system dynamics
\begin{equation}
\label{eq:transition}
   \bm{x}_{t+1} = f(\bm{x}_t, \bm{u}_t),
\end{equation}
produces the trajectory \mbox{$(\bm{x}_0, \bm{u}_0, \bm{x}_1, \bm{u}_1, \ldots, \bm{x}_T)$}. We define the finite-horizon cost as
\begin{equation}
        S(\bm{U};\bm{x}_0) = - \sum_{t=0}^{T-1} r(\bm{x}_t, \bm{u}_t) - r(\bm{x}_T),
\end{equation}
where $T$ is the task horizon and $r(\bm{x}_T)$ the terminal reward. The objective is to find an action sequence that minimizes this finite-horizon cost.

Sampling-based methods address this problem through repeated forward evaluations of candidate action sequences. The computational cost of the dynamics therefore directly constrains the number of trajectories that can be explored within a given computation budget. 

\subsection{Deformable Dynamics}
\label{subsec:deformable_dynamics}

The benchmark requires a dynamics model that can reproduce sufficiently detailed deformation, elastic responses, and contact interactions while remaining inexpensive enough for repeated rollout evaluation. We therefore use reduced Cosserat-rod models, which capture distributed deformation and dynamic elastic behavior with substantially fewer degrees of freedom than general volumetric discretizations. This compromise is particularly relevant for sampling-based control, where each optimization requires many evaluations of the dynamics. The control methods themselves are not specific to Cosserat models and require only forward simulation of the underlying dynamics. 

A rod is parameterized by a material coordinate $X \in [0, L]$ and by time $t$. Each cross-section carries a rigid transform
\begin{equation}
    g(X, t)  \in SE(3),
\end{equation}
which describes both the position of the centerline and the orientation of the local material frame. 

The local deformation of the rod is described by the body strain 
\begin{equation}
\label{eq:cosserat_strain}
    \bm{\xi}(X, t) = \left( g(X,t)^{-1} \partial_X g(X, t)\right)^\vee \in \mathbb{R}^6,
\end{equation}
and the body velocity by 
\begin{equation}
\label{eq:cosserat_velocity}
\bm{\eta}(X, t) = \left(g(X, t)^{-1} \partial_t g(X, t)\right)^\vee  \in \mathbb{R}^6.
\end{equation}
With the angular-first convention, $\bm{\xi}$ contains bending/twist and shear/extension components, while $\bm{\eta}$ contains angular and linear velocity components expressed in the body frame. 

To obtain a finite-dimensional model, the strain field is approximated with a reduced basis
\begin{equation}
\label{eq:reduced_strain}
    \bm{\xi}(X, t) = \bm{\xi}_0(X) +  \bm{\Phi}(X)\bm{q}(t),
\end{equation}
where $\bm{\xi}_0$ is the rest strain, $\bm{\Phi}$ is a matrix of strain modes, and $\bm{q}(t)$ are generalized strain coordinates. Different choices of $\bm{\Phi}$ lead to different reduced strain parameterizations. For example, piecewise-constant strain bases assign one strain vector to each rod segment. In our benchmark, the choice of strain modes is part of the robot model and determines the dimension of the generalized coordinates, the evaluation of the strain at quadrature points, and the trade-off between accuracy and computational cost. 

The pose field is reconstructed from the strain field by integrating the spatial reconstruction equation 
\begin{equation}
\partial_X g(X,t) = g(X,t)\bm{\xi}(X,t)^\wedge.
\end{equation}

The dynamics are obtained by balancing internal elastic wrenches, external wrenches, actuation, and inertia. In compact form, the Cosserat momentum balance can be written as
\begin{equation}
    \label{eq:cosserat_balance}
    \partial_X \bm{\Lambda} - \mathrm{ad}_{\bm{\xi}}^\top \bm{\Lambda} + \bar{\mathbf{F}}_{\mathrm{ext}} = \partial_t (M\bm{\eta}) - \mathrm{ad}_{\bm{\eta}}^\top (M\bm{\eta}),
\end{equation}
where $\bm{\Lambda}$ is the internal wrench, $M$ is the inertia operator per unit length, and $\bar{\mathbf{F}}_{\mathrm{ext}}$ is an external wrench density. A constitutive law closes the system by relating the internal wrench to the elastic strain, for example through
\begin{equation}
   \bm{\Lambda} = \frac{\partial \Psi}{\partial \bm{\epsilon}}, \qquad \bm{\epsilon} = \bm{\xi} - \bm{\xi}_0,
\end{equation}
for some energy density $\Psi$.

Several computational methods compute the reduced dynamics. In the Jacobian-based approach, reduced strain velocities and accelerations are explicitly mapped to body twists and inertial quantities through kinematic Jacobians, and the forces are then projected back to generalized coordinates. In the Newton--Euler recursive approach, the velocities and accelerations are propagated along the rod, local inertial wrenches are computed, and the wrenches are propagated backward before projecting them onto the strain basis. These two formulations are equivalent and yield the same reduced dynamics, but rely on different computational implementations~\cite{ouyoucef2025duality}.

In practice, the continuous rod is discretized into elements and evaluation points. Strains, material laws, internal forces, external forces, actuation, inertia, and contacts are assembled into generalized residuals. A time step is obtained by solving the discrete dynamics residual and applying a time integrator. From the point of view of the controller, these modeling and discretization choices are encapsulated in the transition map~\eqref{eq:transition}. For the remainder of the paper, all sampling strategies interact with the deformable system through forward evaluations of this map. 

\subsection{Sampling-Based Control}

Sampling-based methods search for an action sequence through repeated forward simulation, without requiring derivatives of the full nonlinear dynamics or explicitly enforcing them as equality constraints. Their main limitation is the number of forward simulations required, which motivates the low-sample comparison considered in this work.

During one refinement iteration, a batch of $N$ candidate action sequences is generated according to the current sampling strategy and action parameterization. Denoting by $\bm{\theta}_i$ the parameters associated with candidate $i$, the corresponding action sequence $\bm{U}_i = \pi_{\bm{\theta}_i}$ is rolled out through the simulator, and assigned the cost
\begin{equation}
    S_i = S\left(\pi_{\bm{\theta}_i};\bm{x}_0\right).
\end{equation}

The resulting batch $\{(\bm{\theta}_i,S_i)\}_{i=1}^N$ is used according to the sampling strategy to select a candidate or construct the next sampling distribution. Multiple refinement iterations may be performed successively. Throughout the paper, the rollout budget is defined as
\begin{equation}
    B = NR,
\end{equation}
where $N$ is the number of candidate trajectories per refinement and $R$ is the number of refinement iterations.

\subsection{Action Parameterization}

The parameter vector $\bm{\theta}$ determines how a candidate action sequence is represented. The complete action sequence may either be sampled directly or generated from a lower-dimensional set of parameters.  

The raw parameterization directly uses the action sequence as the search variable:
\begin{equation}
    \bm{U} =\bm{\theta} .
\end{equation}
This is the most general representation, but its dimension grows with both the horizon and the action dimension. Independent perturbations may also produce high-frequency action sequences that are poorly suited to some deformable control tasks. 

The spline parameterization instead represents the action sequence through a lower-dimensional vector of control points $\bm{\theta}$, then reconstructs the full sequence by interpolation:
\begin{equation}
    \bm{U} = \pi_{\bm{\theta}}.
\end{equation}
This reduces the search-space dimension and biases the search toward temporally smoother action sequences. For example, $\pi_{\bm{\theta}}$ can be a linear spline map or a cubic Hermite spline.
The parameterization is independent of the sampling strategy. This allows the comparative analysis to distinguish the effect of temporal action structure from the information used to guide sampling.

\subsection{Sampling Strategies}

We organize the considered strategies according to the information used to guide the search. The presentation starts from direct trajectory sampling and selection, then introduces methods that adapt the sampling distribution from the evaluated population, methods that continuously reweight trajectories according to their costs, and finally a locally model-informed proposal. The formulations are recalled within a common notation, complementing
existing overviews of zero-order techniques for
robotics~\cite{Armand2025ZeroOrder}, to make explicit what additional information is introduced at each level. This organization reflects increasing sources of information and computational complexity, rather than a hierarchy of performance. 

In principle, candidate parameters can be drawn from any sampling distribution. In practice, the distribution must remain sufficiently simple to sample and update repeatedly within a limited rollout budget. We therefore consider Gaussian search distributions throughout the benchmark, 
\begin{equation}
\label{eq:gaussian_parameters}
    p(\bm{\theta}) = \mathcal{N}(\bar{\bm{\theta}}, \Sigma),
\end{equation}
where $\bar{\bm{\theta}}$ defines the nominal parameter vector and $\Sigma$ controls the scale and correlations of the exploration around it. This representation provides a tractable search family whose mean and covariance can either remain fixed or be adapted using the information available to each method.

\subsubsection{Trajectory Sampling and Selection}

The least informed strategies use forward rollouts only to select a candidate trajectory, without estimating the sampling distribution from the complete evaluated population.

\paragraph{Random Shooting}

The most direct use of the rollout budget is to draw every candidate independently from a fixed distribution,
\begin{equation}
    \bm{\theta}_i \sim p_0, \qquad i=1,\ldots,N,
\end{equation}
and returns the lowest-cost candidate:
\begin{equation}
    i^\star = \arg\min_i S_i, \qquad \bm{\theta}_{\mathrm{new}} = \bm{\theta}_{i^\star}.
\end{equation}

No information from the sampled batch is used to update $p_0$, making Random Shooting the reference case in which the rollout budget is used without adapting the search.

\paragraph{Predictive Sampling~\cite{Howell2022PredictiveSR}}

Best-sample selection can be warm-started by centering each batch on the candidate selected at the previous refinement. Candidates are sampled as
\begin{equation}
    \bm{\theta}_i \sim \mathcal{N}(\bar{\bm{\theta}}, \Sigma).
\end{equation}
and the nominal parameter is updated to the best sampled candidate:
\begin{equation}
    i^\star = \arg\min_i S_i, \qquad \bar{\bm{\theta}} \leftarrow \bm{\theta}_{i^\star}.
\end{equation}

Each subsequent batch is therefore centered on the best candidate from the previous refinement. In receding-horizon MPC, the selected sequence is additionally shifted before the next planning step. Predictive Sampling therefore preserves the simplicity of best-sample selection while warm-starting both successive refinements and successive MPC updates.

\subsubsection{Distribution Adaptation}

Rather than updating the nominal parameter from a single selected candidate, distribution-adaptive methods use information from several evaluated trajectories to update the search distribution. This allows the distribution itself to evolve across successive refinements. 

\paragraph{Cross-Entropy Method~\cite{rubinstein1999cross}}

Rather than selecting a single candidate, the search distribution itself can be refit to the best fraction of the batch. At iteration $k$, candidates are sampled from
\begin{equation}
    p_k(\bm{\theta}) = \mathcal{N}(\bar{\bm{\theta}}_k, \Sigma_k).
\end{equation}

After rollout evaluation, the method keeps the elite set
\begin{equation}
    \mathcal{E}_k = \left\{ i \;:\; S_i \leq Q_\eta(S_1, \ldots, S_N) \right\},
\end{equation}
where $Q_\eta$ is the empirical $\eta$-quantile of the sampled costs. The elite samples define
\begin{align}
    \bm{\mu}_{\mathcal{E}} &= \frac{1}{|\mathcal{E}_k|}\sum_{i\in\mathcal{E}_k} \bm{\theta}_i, \\
    \Sigma_{\mathcal{E}} &= \frac{1}{|\mathcal{E}_k|} \sum_{i\in\mathcal{E}_k}(\bm{\theta}_i-\bm{\mu}_{\mathcal{E}})(\bm{\theta}_i-\bm{\mu}_{\mathcal{E}})^\top.
\end{align}

The search distribution is updated with an exponential moving average
\begin{align}
    \bar{\bm{\theta}}_{k+1}  &= (1 - \gamma)\bar{\bm{\theta}}_k + \gamma\bm{\mu}_{\mathcal{E}}, \\
    \Sigma_{k+1} &= (1-\gamma)\Sigma_k + \gamma\Sigma_{\mathcal{E}}.
\end{align}

In our implementation, $\Sigma_k$ is diagonal, so the mean and variance are updated component-wise from the elite samples. This remains a hard-threshold update: samples outside the elite set do not contribute.

\paragraph{CMA-ES~\cite{hansenCMAES2001}}

A Gaussian search distribution can also be adapted through ranking, recombination, and covariance evolution rather than an elite refit. Candidates are sampled as
\begin{equation}
    \bm{\theta}_i = \bar{\bm{\theta}}_k + \sigma_k \bm{y}_i, \qquad \bm{y}_i \sim \mathcal{N}(\bm{0}, C_k),
\end{equation}
where $C_k$ is the covariance matrix and $\sigma_k$ is a global step size. 
After ranking candidates by cost, the mean is recombined from the best samples: 
\begin{equation}
    \bar{\bm{\theta}}_{k+1} = \sum_{i=1}^{\mu}a_i \bm{\theta}_{i:N},
\end{equation}
where $\mu \leq N$ is the number of selected parents, $\bm{\theta}_{i:N}$ denotes the $i$-th best sample after ranking the $N$ candidates by increasing cost, and $a_i$ are recombination weights. 

The covariance and global step size are then updated through the standard CMA-ES evolution paths. CMA-ES thereby adapts both the global exploration scale and correlations between search directions.

\subsubsection{Cost-Weighted Sampling}

The previous strategies either select individual trajectories or estimate a new search distribution from population statistics. Path-integral methods instead use the rollout cost of every sampled trajectory to continuously weight its contribution to the update. The MPPI weighting rule can be motivated through a regularized optimization over probability distributions.

Let $p_0(\bm{\theta})$ denote the nominal sampling distribution. We seek a distribution $p$ that minimizes the expected rollout cost while remaining close to $p_0$:
\begin{equation}
    \label{eq:kl_control}
    \min_p\; \mathbb{E}_{\bm{\theta}\sim p}[S(\pi_{\bm{\theta}};\bm{x}_0)] + \lambda\mathrm{KL}(p\,\|\,p_0),
\end{equation}
where $\lambda>0$ controls the trade-off between rollout cost and deviation from the nominal distribution.

The optimal distribution has the Boltzmann form
\begin{equation}
\label{eq:boltzmann_distribution}
    p^\star(\bm{\theta}) = \frac{1}{Z} p_0(\bm{\theta})\exp\left( -\frac{S(\pi_{\bm{\theta}};\bm{x}_0)}{\lambda}\right),
\end{equation}
where $Z$ is a normalization constant. The nominal distribution therefore acts as a prior over the search, while the exponential cost term assigns greater probability mass to lower-cost trajectories.

\paragraph{Vanilla MPPI~\cite{Williams2017InformationTheoreticMP}} 

The cost-weighted distribution in~\eqref{eq:boltzmann_distribution} is approximated on a finite batch sampled from the nominal Gaussian distribution $p_0$.

The continuous reweighting is represented on the finite batch by normalized discrete weights
\begin{equation}
    w_i = \frac{\exp\left(-(S_i-\rho)/\lambda\right)}{\sum_j \exp\left(-(S_j-\rho)/\lambda\right)}, \qquad \rho = \min_j S_j,
\end{equation}
where $\rho$ improves numerical stability. 
The nominal parameter is updated as the weighted mean of the sampled candidates:
\begin{equation}
    \bar{\bm{\theta}} \leftarrow \sum_i w_i \bm{\theta}_i.
\end{equation}

With fixed covariance, this weighted mean moves the nominal parameter toward regions represented by lower-cost samples. The temperature $\lambda$ controls how strongly the update concentrates on the lowest-cost trajectories. Normalizing the rollout costs makes the temperature less sensitive to the absolute scale of the task cost.

The following variants retain this cost-weighted update while modifying the exploration schedule, proposal distribution, or covariance of the search.

\paragraph{Annealed MPPI~\cite{xue2024dialmppi,schramm2025reference}}

The exploration scale can additionally be scheduled across refinement iterations and control points. For spline parameterizations, the scale at refinement iteration $r$ and control-point index $k$ is
\begin{equation}
    \sigma_{r,k} = \sigma_0 \exp \left(-\kappa\frac{r}{R} - \beta\frac{K-1-k}{K} \right),
\end{equation}
where $R$ is the number of refinement iterations and $K$ is the number of control points. At refinement step $r$, candidates are sampled as
\begin{equation}
    \bm{\theta}_{i,r} = \bar{\bm{\theta}}_r + \bm{\epsilon}_{i,r},
\end{equation}
with 
\begin{equation}
    \epsilon_{i,r,k} = \sigma_{r,k} z_{i,r,k}, \qquad z_{i,r,k} \sim \mathcal{N}(0,1).
\end{equation}

This schedule combines coarse-to-fine exploration across refinements with smaller perturbations on control points closer to execution.

\paragraph{Importance-Sampling MPPI}

The proposal used to generate candidate trajectories need not coincide with the nominal distribution in~\eqref{eq:boltzmann_distribution}. Candidates are drawn from a proposal $q$, and their weights are corrected by the density ratio $p_0(\bm{\theta}_i)/q(\bm{\theta}_i)$. The unnormalized log-weights are
\begin{equation}
    \log \tilde{w}_i = - \frac{S_i -\rho}{\lambda} + \log p_0(\bm{\theta}_i) - \log q(\bm{\theta}_i),
\end{equation}
and the normalized weights are
\begin{equation}
    w_i = \frac{\exp(\log \tilde{w}_i)}{\sum_j\exp(\log \tilde{w}_j)}.
\end{equation}

When $q=p_0$, the density-ratio correction cancels and the method reduces to vanilla MPPI.

\paragraph{Covariance-Adapted MPPI}

The same cost weights can also adapt the exploration variance, not only the nominal parameter. After the weighted-mean update, the diagonal variance is estimated from the weighted sample dispersion,
\begin{equation}
    \bm{\sigma}^2 \leftarrow \sum_i w_i \left(\bm{\theta}_i - \bar{\bm{\theta}} \right)^2.
\end{equation}

The estimate is smoothed and clipped to avoid variance collapse or excessive exploration. This reduces exploration in directions where high-weight samples are concentrated and preserves it where useful samples remain dispersed.

\subsubsection{Locally Model-Informed Sampling}

The strategies above guide the search using sampled trajectories, their rollout costs, and statistics or prescribed modifications of the sampling distribution. Model-Guided MPPI~\cite{schramm2026variancereduced} introduces an additional source of information by exploiting local differential information from the dynamics and the smooth components of the task objective around the current nominal trajectory. The nonlinear simulator remains responsible for evaluating every candidate trajectory, while the local model is used to construct a more informative proposal. To separate this local information from the complete rollout evaluation, we write
\begin{equation}
    S(\pi_{\bm{\theta}}; \bm{x}_0)= m(\bm{\theta}) + \Delta(\bm{\theta}),
\end{equation}
where $m$ is a differentiable local approximation and $\Delta$ contains the contribution of the full rollout cost not represented by this approximation.

\paragraph{Local quadratic model}

We approximate $m$ around the current nominal parameter $\bar{\bm{\theta}}$ using
\begin{equation}
    m_{\mathrm{quad}}(\bm{\theta}) = m(\bar{\bm{\theta}}) + \bm{g}^{\top} (\bm{\theta}-\bar{\bm{\theta}}) + \frac{1}{2} (\bm{\theta}-\bar{\bm{\theta}})^{\top} H (\bm{\theta}-\bar{\bm{\theta}}).
\end{equation}

When $m$ is expressed as a least-squares cost over smooth task residuals, 
\begin{equation}
    m(\bm{\theta}) = \sum_{t\in\mathcal{T}_m} \frac{1}{2} \left\| \bm{e}_t(\bm{\theta})\right\|^2,
\end{equation}
the Gauss-Newton approximation gives
\begin{equation}
    \bm{g} = J^\top\bm{e}(\bar{\bm{\theta}}), \qquad H \approx J^\top J,
\end{equation}
where $J= \partial \bm{e} / \partial \bm{\theta}$ is evaluated at $\bar{\bm{\theta}}$. 

The residual vector $\bm{e}$ contains only smooth, differentiable terms, such as tracking, deformation, base-position, and action-regularization terms. Nonsmooth terms, including barriers and contact-dependent costs, are excluded from the local approximation and remain evaluated exclusively through the full nonlinear rollout.

\paragraph{Randomized Gauss-Newton approximation}

In practice, we do not assemble $J$ explicitly. We only require Jacobian-transpose products of the form $J^\top \bm{y}$, computed using a hand-derived adjoint pass through the implicit time step. 

We approximate the Gauss-Newton curvature using randomized Rademacher probes~\cite{HutchinsonEstimator}. Let
\begin{equation}
\bm v_p = J^\top\bm z_p, \qquad \bm z_p\in\{-1,1\}^{d_e},
\end{equation}
where $d_e$ is the dimension of the residual space. The resulting approximation is
\begin{equation}
\widehat H =   \frac1P \sum_{p=1}^P \bm v_p \bm v_p^\top.
\end{equation}

For large parameter spaces, only the diagonal of this approximation is retained,
\begin{equation}
\operatorname{diag}(\widehat H) = \frac1P \sum_{p=1}^P \bm v_p \odot \bm v_p.
\end{equation}

The dense approximation is used for moderate parameter dimensions, whereas only its diagonal is retained for larger parameter spaces in order to reduce computational cost.

\paragraph{Guided proposal}

The influence of the local model is controlled by a weight $\alpha \in [0,1]$. Combining the weighted quadratic model with the prior gives
\begin{equation}
    \tilde{\Sigma} = \left(\Sigma^{-1} + \frac{\alpha}{\lambda}\widehat H \right)^{-1}, \qquad \tilde{\bm{\theta}} = \bar{\bm{\theta}} - \frac{\alpha}{\lambda} \tilde{\Sigma}\bm{g}.
\end{equation}

Samples are drawn from $\bm{\theta}_i \sim \mathcal{N}\left(\tilde{\bm{\theta}}, \tilde{\Sigma} \right)$. In practice, the guided shift is clipped to avoid moving too far from the nominal trajectory, yielding a stabilized approximation of the guided proposal.

\paragraph{Exploration and model activation}

Local guidance need not be active during every refinement. Early refinements may use $\alpha=0$, reducing the proposal to the unguided MPPI distribution and preserving broad exploration. Guidance can then be activated directly or progressively by increasing $\alpha$ once the nominal trajectory reaches a sufficiently informative regime. 

The set $\mathcal{T}_m$ determines the portion of the trajectory on which the local model is constructed. It may cover the complete horizon or only a selected phase, and remains fixed throughout a refinement. Before the next refinement, however, $\mathcal{T}_m$ can be updated from information gathered in the previous batch. For example, an event detected on the updated nominal trajectory, such as the onset of contact, can be used to restrict the local model to the corresponding phase of the trajectory.

\paragraph{Residual weighting}

Since the weighted local quadratic model has already been incorporated into the guided proposal, the remaining contribution required to recover the full rollout weighting is
\begin{equation}
    \tilde{\Delta}_i = S(\pi_{\bm{\theta}_i};\bm{x}_0) - \alpha m_{\mathrm{quad}}(\bm{\theta}_i).
\end{equation}

The corresponding MPPI weights are
\begin{equation}
    w_i = \frac{\exp\left(-(\tilde{\Delta}_i-\tilde{\rho})/\lambda\right)}{\sum_j\exp\left(-(\tilde{\Delta}_j-\tilde{\rho})/\lambda\right)}, \qquad \tilde{\rho} = \min_j \tilde{\Delta}_j.
\end{equation}

When $\alpha=1$ and the quadratic approximation is accurate, $\tilde{\Delta}_i$ approximates the residual $\Delta(\bm{\theta}_i)$. For $0<\alpha<1$, it retains the portion of the local model that was not absorbed into the proposal. When $\alpha=0$, the proposal and weighting reduce to vanilla MPPI.

The nominal parameter is finally updated using the standard MPPI weighted mean. Model guidance therefore modifies both the proposal and the residual cost used for weighting, while the full nonlinear rollout remains the reference for evaluating every sampled trajectory.

\section{Comparative Analysis}
\label{sec:experiments}

We evaluate sampling-based control along four complementary axes: (i) open-loop sample efficiency, (ii) dynamic structure of the optimized trajectories, (iii) receding-horizon control, and (iv) computational performance. The benchmark comprises the five environments illustrated in~\cref{fig:benchmark_task}, spanning deformable stabilization, contact-mediated reaching, passive-object interaction, elastic energy storage, and higher-dimensional manipulation with a deformable end-effector. The accompanying video shows representative open-loop and closed-loop executions for all environments.

\subsection{Experimental setup}
\label{subsec:experimental_setup}

\noindent\textbf{Simulation}. The deformable systems follow the Cosserat rod formulation introduced in~\Cref{subsec:deformable_dynamics} and use a reduced-strain implementation in the same modeling spirit as SoRoSim~\cite{AnupSorosim2023}. Contact is represented by a spring-damper penalty model, in which penetration generates repulsive forces and velocity-dependent damping stabilizes the contact response. This model is deliberately simple, since the benchmark aims to evaluate sampling-based control under deformable, contact-dependent dynamics.

The simulator is implemented in JAX and supports batched rollouts and just-in-time compilation. Candidate action sequences are evaluated in parallel using the same compiled dynamics implementation on CPU and GPU. The timing experiments use direct JAX compilation of the simulator for the corresponding hardware, without architecture-specific GPU kernels. We will release the implementation upon paper acceptance.

\noindent\textbf{Evaluation protocol}. Within each environment, we evaluate all methods using the same dynamics, task cost, initial conditions, action bounds, horizon, and rollout budget. Unless stated otherwise, we report quantitative comparisons over five random seeds. We select hyperparameters from coarse predefined grids for each method and environment. Comparisons under equal rollout budget use the best configuration found within these grids, whereas fixed-seed executions are reported only to illustrate the existence and structure of successful trajectories.

In the open-loop setting, each method optimizes one complete action sequence from a fixed initial state. In the receding-horizon setting, each method optimizes a sequence of $H$ actions, executes only the first action, and replans from the resulting state.

\begin{figure}[h]
    \centering
    \includegraphics[width=0.5\textwidth]{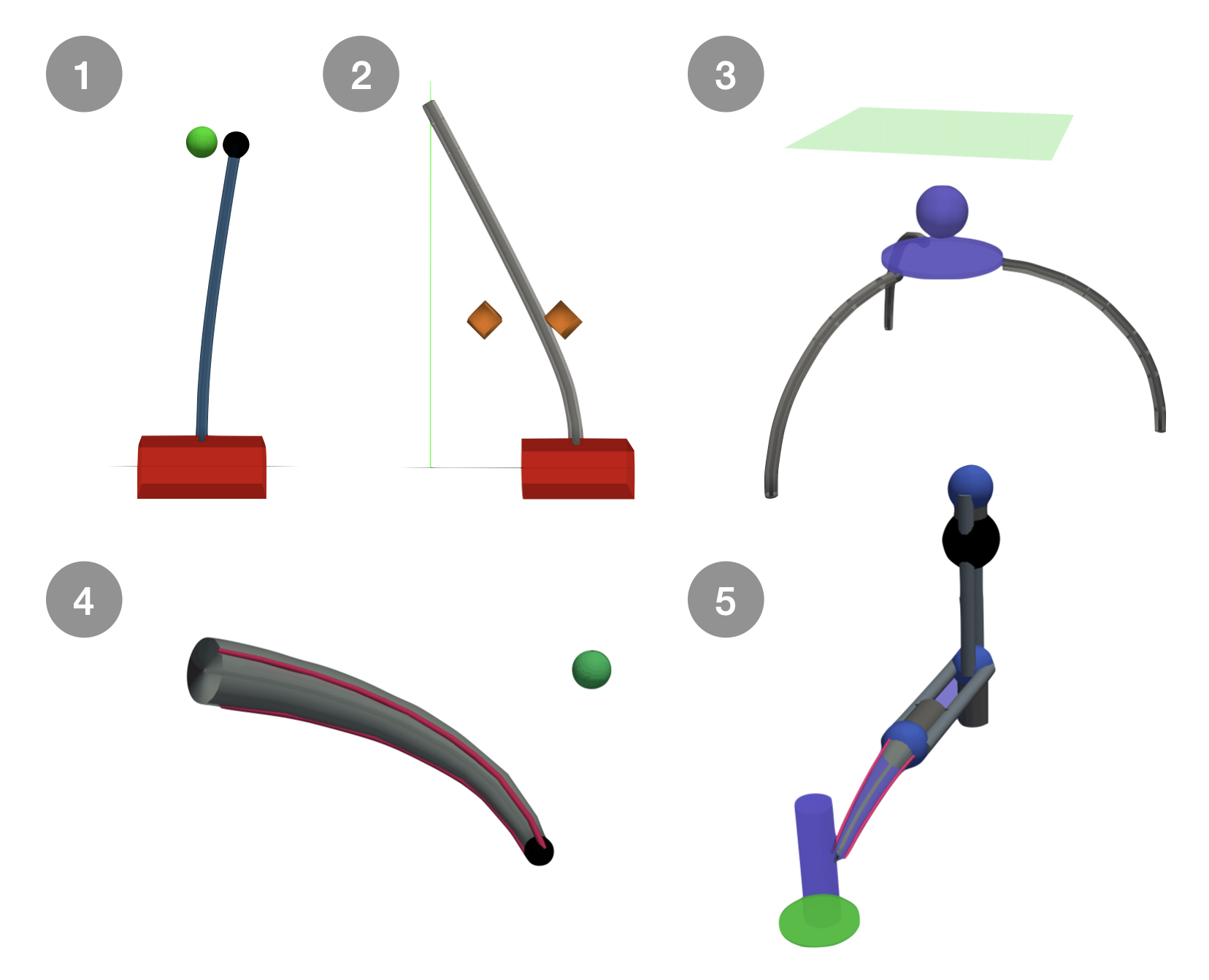}
    \caption{\textbf{Benchmark environments}. From left to right: 1) a deformable Cartpole, 2) a Cart-Stem with obstacles, 3) a Tripod robot interacting with a sphere, 4) a Deformable Trunk, and 5) a Three-axis rigid robot with a Trunk effector. Task targets are represented in green.}
    \label{fig:benchmark_task}
\end{figure}

\begin{figure*}[!h]
    \centering
    \includegraphics[width=0.85\textwidth]{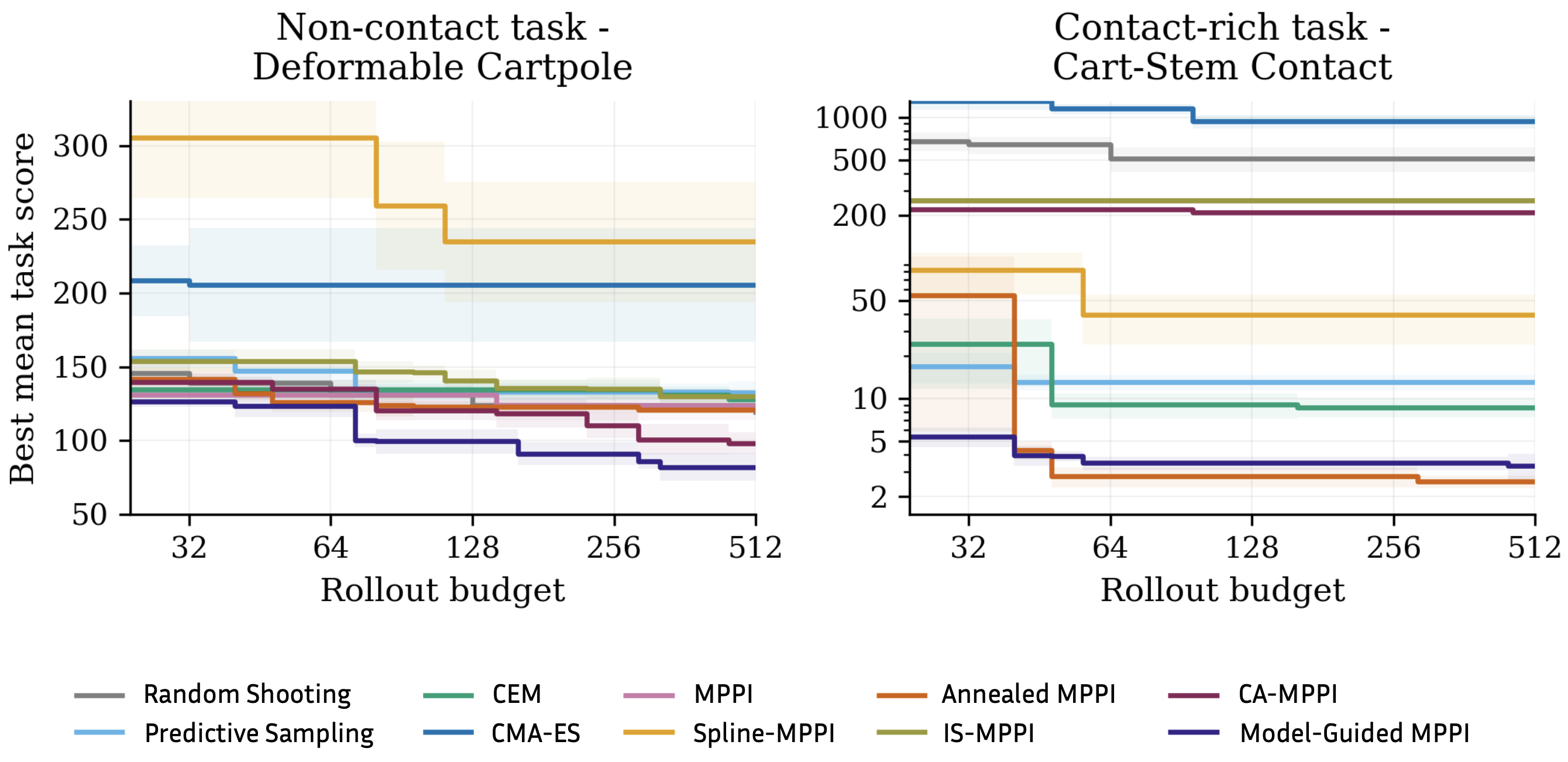}
    \caption{\textbf{Open-loop optimization under a fixed rollout-budget}. The reported value at budget $B$ is the best mean task score obtained using at most $B$ rollouts. Curves and shaded regions show the mean and standard error of the mean over five seeds. The Cartpole uses a dense non-contact objective, whereas the Cart-Stem requires discovering a useful contact mode. Lower scores are better.}
    \label{fig:open_loop_equal_budget}
\end{figure*}

\noindent\textbf{Benchmark environments}. The benchmark combines five environments covering complementary control challenges.

\begin{itemize}
    \item \textit{Deformable Cartpole} (\cref{fig:benchmark_task}.1) is adapted from the classical Cartpole problem by replacing the rigid pole with a Cosserat rod. The objective is to stabilize the flexible pole while keeping the cart near the rail center. Success is declared when the terminal tip error is below $25$~mm.
    \item \textit{Cart-Stem Contact} (\cref{fig:benchmark_task}.2) is adapted from~\cite{menager2024learning}. The controller must exploit obstacle contact before reaching the target. Success requires a terminal tip error below $5$~mm together with a cart velocity below $10$~mm/s.
    \item \textit{Tripod-Sphere} (\cref{fig:benchmark_task}.3) is inspired by~\cite{menager2025frictional}. Three deformable legs must cooperatively launch a passive sphere above $0.218$~m. Success requires a terminal sphere-to-target error below $1$~mm.
    \item \textit{Dynamic Trunk} (\cref{fig:benchmark_task}.4) requires exploiting elastic energy to reach a target that is unreachable through direct cable pulling. Success corresponds to a terminal tip error below $12$~mm.
    \item \textit{Three-axis Trunk} (\cref{fig:benchmark_task}.5) is the largest manipulation task. A deformable trunk mounted on a rigid positioning stage must push a cylinder to a target through contact. Success requires a terminal planar error below $18$~mm.
\end{itemize}

Across all environments, the rollout objective combines task-specific terms with regularization on motion, deformation, and actuation. Success criteria are defined as task-specific tolerances chosen for evaluation only and do not enter the optimization objective.

\begin{table*}[!h]
  \caption{Representative successful open-loop configurations at a fixed seed. Hyperparameters are selected through coarse grid search. The table demonstrates task feasibility and is not a statistical comparison between methods.}
  \label{tab:sol_open_loop}
  \centering
  \begin{adjustbox}{max width= \textwidth}
    \begin{tabular}{r|ccccc}
      \toprule
      \textbf{Task} & \textbf{Terminal task error} (mm) & \textbf{Threshold} (mm) & \textbf{Method} & \textbf{Optimization horizon} & \textbf{Rollout budget}\\
      \midrule
      \rowcolor{pastelblue}
      Cartpole & 11.6 & 25.0 & Model-Guided MPPI & 500 & 320 \\
      Cart-Stem & 3.4 & 5.0 & Model-Guided MPPI & 200 & 448 \\
      \rowcolor{pastelblue}
      Dynamic Trunk &  7.2 & 12.0 & CEM & 33 & 224 \\
      Tripod-Sphere & 0.04 & 1.00 & Predictive Sampling & 50 & 80\\
      \rowcolor{pastelblue}
      Three-axis Trunk & 4.3 & 18.0 & Model-Guided MPPI & 400 & 96 \\
      \bottomrule
    \end{tabular}
  \end{adjustbox}
\end{table*}

\subsection{Open-loop sample efficiency}
\label{subsec:open_loop}

We first compare the sampling strategies under matched rollout budgets in two complementary environments: the Deformable Cartpole, providing a dense non-contact objective, and the Cart-Stem, requiring the discovery and exploitation of contact. This open-loop setting isolates sample efficiency from the effects of replanning.

The equal-budget comparison in~\cref{fig:open_loop_equal_budget} shows a difference between the two tasks. On the Deformable Cartpole, several methods achieve comparable performance at low rollout budgets because the objective provides informative feedback throughout the trajectory and the relevant behaviors remain within a relatively coherent mode. Raw action parameterization is also effective in this environment. The optimized controls use rapid corrective oscillations to maintain the flexible pole upright, whereas a low-dimensional spline basis restricts part of this high-frequency control authority. This behavior is specific to the Cartpole stabilization task considered here. 

The Cart-Stem produces a substantially stronger separation between methods. Its solution requires a temporally coordinated motion that first establishes obstacle contact and then exploits this configuration to approach the target. In the tested low-budget setting, spline parameterizations are advantageous because they reduce the search dimension and bias the sampled controls toward temporally coherent motions. With 24 rollouts, Model-Guided MPPI is the only method that succeeds across all five seeds. At approximately twice this budget, Annealed MPPI reaches comparable performance, while Predictive Sampling and CEM also succeed consistently but retain higher terminal costs. The main difficulty is therefore not only selecting low-cost trajectories, but discovering a useful contact mode before refining the solution within that mode.

This distinction explains the poor behavior of the finite-batch weighted-mean MPPI update on the contact task. When several qualitatively different strategies receive high weight, their average does not necessarily remain within any successful mode. For example, trajectories that establish different contact configurations, or trajectories that respectively create and avoid contact, may average into a sequence that realizes neither behavior. Predictive Sampling retains an evaluated trajectory, while CEM refits its distribution using an elite subset. These updates preserve coherent sampled strategies more directly and are therefore less vulnerable to averaging across incompatible contact modes.

Model guidance is most effective after sampling has identified an informative local regime. In the Cart-Stem experiments, the first three refinements are unguided and are used to discover a contact configuration. The local model is activated only afterward to refine the trajectory within the selected mode. Reducing this exploratory phase to one refinement degrades performance. The local model improves sample efficiency inside an already discovered mode. The transition into that mode remains a sampling problem. 

The remaining performance differences reflect the structure each tested implementation can infer from a small number of rollouts. In the implementation evaluated here, Importance-Sampling (IS) MPPI uses a broader Gaussian proposal that introduces neither task-dependent directions nor temporal correlations. The density correction can therefore increase weight concentration without systematically increasing the probability of sampling a useful mode. Covariance-Adapted (CA) MPPI estimates only component-wise variances and cannot represent temporal correlations or coordinated perturbations across actuators. CMA-ES adapts a richer search distribution, but the number of quantities to estimate becomes large relative to the available population when optimizing long action sequences. CEM receives only a few elite samples at the smallest batch sizes, making its distribution update noisy, while Annealed MPPI remains sensitive to the scale of its prescribed exploration schedule. These observations characterize the low-budget implementations evaluated here rather than intrinsic limitations of the corresponding method families.

The controlled comparison above focuses on two environments for which all methods are evaluated under matched rollout budgets. To establish feasibility over the complete benchmark,~\cref{tab:sol_open_loop} reports one representative successful open-loop configuration for each environment. These fixed-seed results illustrate task solvability and are not used for statistical ranking.

\subsection{Dynamic structure of the optimized trajectories}
\label{subsec:dynamic_structure}

The previous comparison evaluates optimization performance but does not explain how successful trajectories exploit deformable dynamics. Across the contact and elastic tasks, successful trajectories share a common structure: the controller temporarily sacrifices the immediate task metric to create a mechanical state from which the terminal objective becomes reachable.

\begin{figure}[h]
    \centering
    \includegraphics[width=0.5\textwidth]{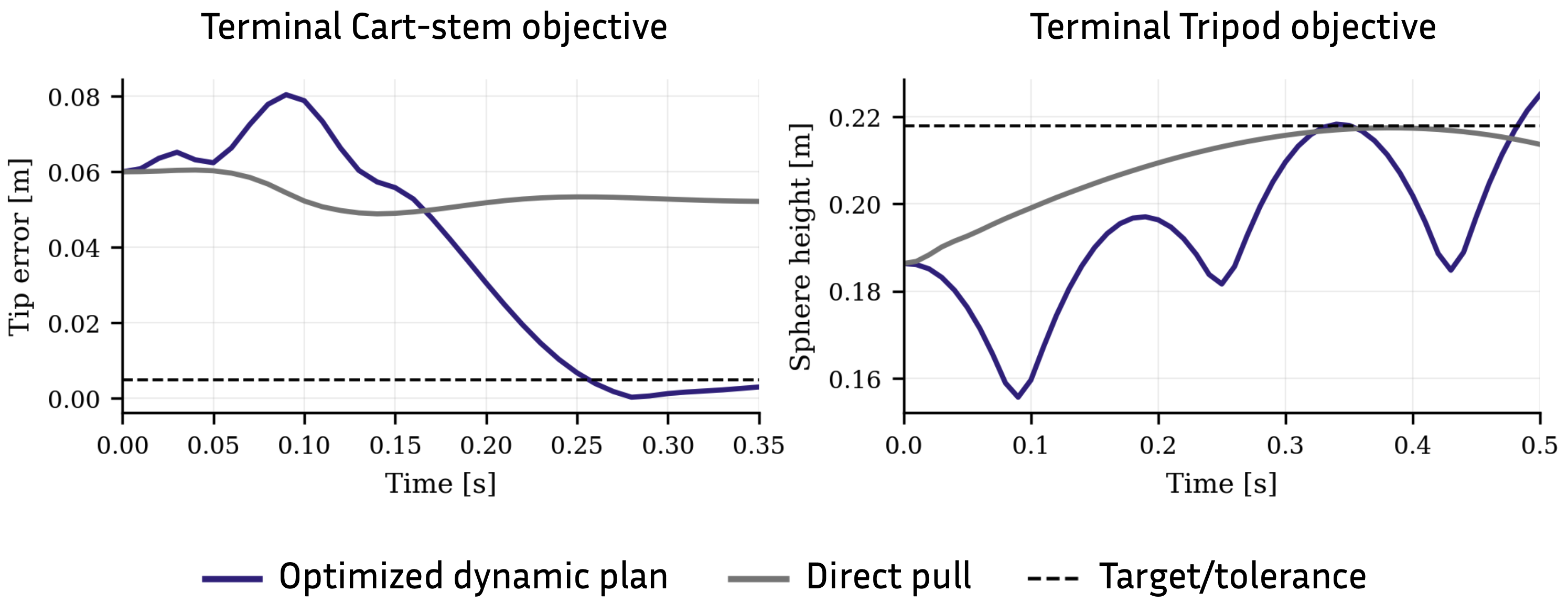}
    \caption{\textbf{Non-monotonic task progress in representative optimized trajectories}. Left: Cart-Stem tip error. The optimized trajectory first moves away from the target to establish obstacle contact before converging to the goal. Right: sphere height in the Tripod-Sphere task. The optimized trajectory first lowers the sphere-platform system before producing a jump. Direct-action heuristics are shown for qualitative comparison. Dashed lines indicate the task target or tolerance.}
    \label{fig:delayed_reward}
\end{figure}

This behavior appears in two environments shown in~\cref{fig:delayed_reward}. In the Cart-Stem task, the optimized trajectory initially increases the tip error while moving the stem toward a useful obstacle contact. Once this contact mode has been established, the error decreases rapidly and crosses the success threshold. The direct-action heuristic decreases the error locally but converges to an unfavorable configuration. 

The Tripod-Sphere task exhibits the same non-monotonic behavior. The controller first lowers the sphere together with the supporting platform, then reverses the platform motion, generating sphere-platform separation and a sequence of upward jumps. The direct upward command produces a largely monotonic lift but does not realize the same motion. In both tasks, terminal success depends on an intermediate motion that is unfavorable under the instantaneous task metric. 

\begin{figure}[h]
    \centering
    \includegraphics[width=0.44\textwidth]{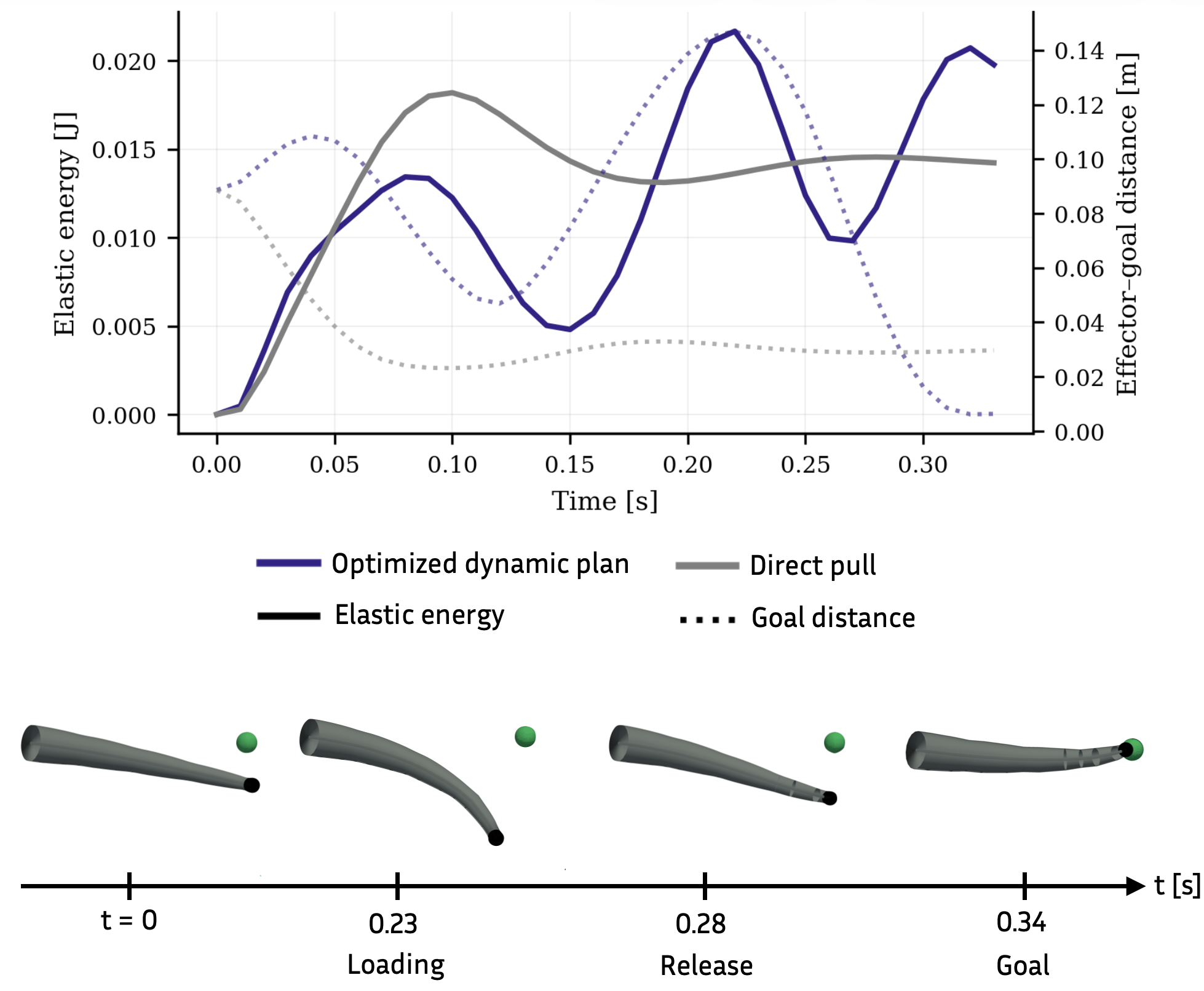}
    \caption{\textbf{Elastic loading and release in the dynamic trunk}. The optimized sequence first bends the trunk away from the target, increasing its elastic energy, and subsequently releases this deformation while actuating toward the goal. The effort-matched direct pull generates elastic deformation but does not realize the coordinated loading-and-release sequence and does not reach the target. Top: Elastic energy and tip-goal distance along the trajectory. Bottom: Representative loading and release sequence.}
    \label{fig:trunk_energy}
\end{figure}

The dynamic trunk makes this mechanism explicit through elastic energy storage and release. In~\cref{fig:trunk_energy} (top), the optimized trajectory first moves the tip away from the target while elastic energy increases. The subsequent decrease in elastic energy coincides with a rapid reduction of the tip-goal distance. The effort-matched constant pull produces deformation but does not coordinate loading and release, and consequently remains at a larger terminal distance. The optimized trajectory therefore exploits the timing of elastic deformation rather than merely applying a larger control effort. This loading-and-release pattern is also directly visible in the representative trajectory shown in Fig.~\ref{fig:trunk_energy} (bottom).

These behaviors also clarify the role of action parameterization. When useful motion depends on coordinated loading, contact, and release phases, sampling independent actions at every time step allocates most of the rollout budget to temporally incoherent sequences. 

The Three-axis Trunk provides the clearest test because its horizon of 400 steps yields 2400 raw action variables. Replacing raw actions with seven spline knots reduces the search dimension to 42. Comparing raw actions with spline parameters therefore combines dimensionality reduction and temporal regularization. Comparing piecewise-constant and cubic parameterizations at the same dimension isolates the effect of temporal continuity.

As reported in~\cref{tab:actuation_param}, raw-action optimization fails to produce useful object displacement within the available $96$-rollout budget. Both reduced parameterizations improve the terminal position, showing the importance of reducing the temporal search dimension. The comparison between the two $42$-parameter representations then isolates the effect of the interpolation basis. The cubic representation reduces the action-variation metric by approximately a factor of $85$ relative to the piecewise-constant representation and produces a substantially smaller final object speed. It therefore imposes a stronger temporal regularity bias while retaining enough freedom to realize the required contact motion.

\begin{table}[!h]
  \caption{Results obtained for the Three-axis Trunk task for different action parameterizations. All results are means over five seeds, with a rollout budget of 96 with Annealed MPPI. Action variation is measured as the horizon-averaged first difference, $\frac{1}{H-1}\sum_{t=0}^{H-2}\|\bm{u}_{t+1}-\bm{u}_t\|$.}
  \label{tab:actuation_param}
  \centering
  \begin{adjustbox}{max width=\columnwidth}
    \begin{tabular}{r|ccc}
      \toprule
      \textbf{Parameterization} & \textbf{Position error} & \textbf{Final speed}  & \textbf{Mean action variation}\\
      \midrule
      \rowcolor{pastelblue}
      Raw & 85.0 mm & 0.0 mm/s & 0.6775 N  \\
      Piecewise constant & 45.0 mm & 32.1 mm/s & 0.9006 N  \\
      \rowcolor{pastelblue}
      Cubic spline &  39.4 mm & 5.4 mm/s & 0.0106 N \\
      \bottomrule
    \end{tabular}
  \end{adjustbox}
\end{table}

\begin{table*}[ht]
  \caption{Representative successful closed-loop solutions at a fixed seed. The rollout budget is reported per MPC update. Hyperparameters are selected through coarse grid search.}
  \label{tab:sol_closed_loop}
  \centering
  \begin{adjustbox}{max width=\textwidth}
    \begin{tabular}{r|cccc}
      \toprule
      \textbf{Task} & \textbf{Terminal task error} [mm] & \textbf{Threshold} [mm] & \textbf{Planning horizon} & \textbf{Rollouts per MPC update}\\
      \midrule
      \rowcolor{pastelblue}
      Cartpole & 0.4 & 25.0 & 16 & 48 \\
      Cart-Stem& 3.0 & 5.0 &  20 & 16 \\
      \rowcolor{pastelblue}
      Dynamic Trunk &  10.0 & 12.0 &  16 & 48 \\
      Tripod-Sphere & 0.03 & 1.00 &  25 & 24 \\
      \rowcolor{pastelblue}
      Three-axis Trunk & 9.0 & 18.0 &  20 & 8 \\
      \bottomrule
    \end{tabular}
  \end{adjustbox}
\end{table*}

\begin{figure*}[!h]
    \centering
    \includegraphics[width=0.8\textwidth]{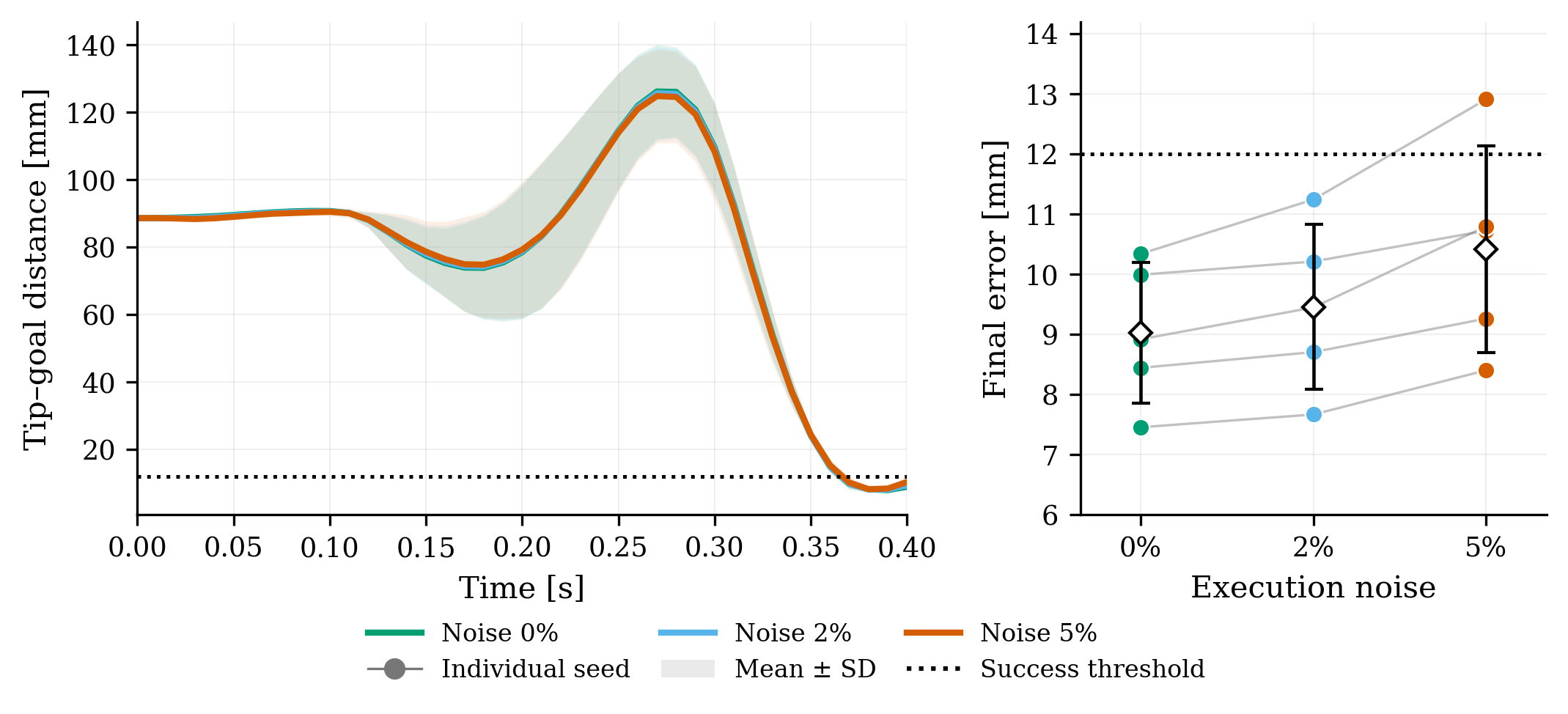}
    \caption{\textbf{Closed-loop robustness of Annealed MPPI on the dynamic trunk under Gaussian execution noise}. Left: mean tip-goal distance with one standard deviation over five seeds. Right: terminal errors for individual seeds, together with the mean and standard deviation. The same controller parameters are used for all noise levels, with $H=16$, $N=16$, and $R=3$.}
    \label{fig:trunk_MPC_robustness}
\end{figure*}

\subsection{Receding-horizon control}
\label{subsec:closed_loop}

The closed-loop experiments assess whether the open-loop observations remain relevant when only the first action is executed and the plan is recomputed from feedback. Annealed MPPI provides the best empirical trade-off between task performance and update cost among the tested methods. Its refinement schedule allocates broad exploration to the first batches and progressively reduces perturbations around the near-term actions, which is well matched to receding-horizon execution. 

Representative successful executions are obtained for all five environments, as summarized in~\cref{tab:sol_closed_loop}. These results establish closed-loop feasibility.

Closed-loop robustness is evaluated on the Dynamic Trunk by applying independent Gaussian perturbations to the executed actions while keeping the controller parameters fixed. Results are shown in ~\cref{fig:trunk_MPC_robustness}. Across five seeds, Annealed MPPI reaches the $12$-mm target in all trials without execution noise and with noise corresponding to $2\%$ of the actuator range. At $5\%$ noise, four of the five trials remain successful, and the mean terminal error increases from $9.03 \pm 1.17$~mm to $10.42 \pm 1.72$~mm. The mean trajectories remain similar up to the release phase, indicating that replanning preserves the loading strategy under moderate disturbances. The increased terminal dispersion at $5\%$ noise is concentrated during the release and final approach, where small actuation errors have the largest effect on the tip trajectory.

\begin{figure}[h]
    \centering
    \includegraphics[width=0.49\textwidth]{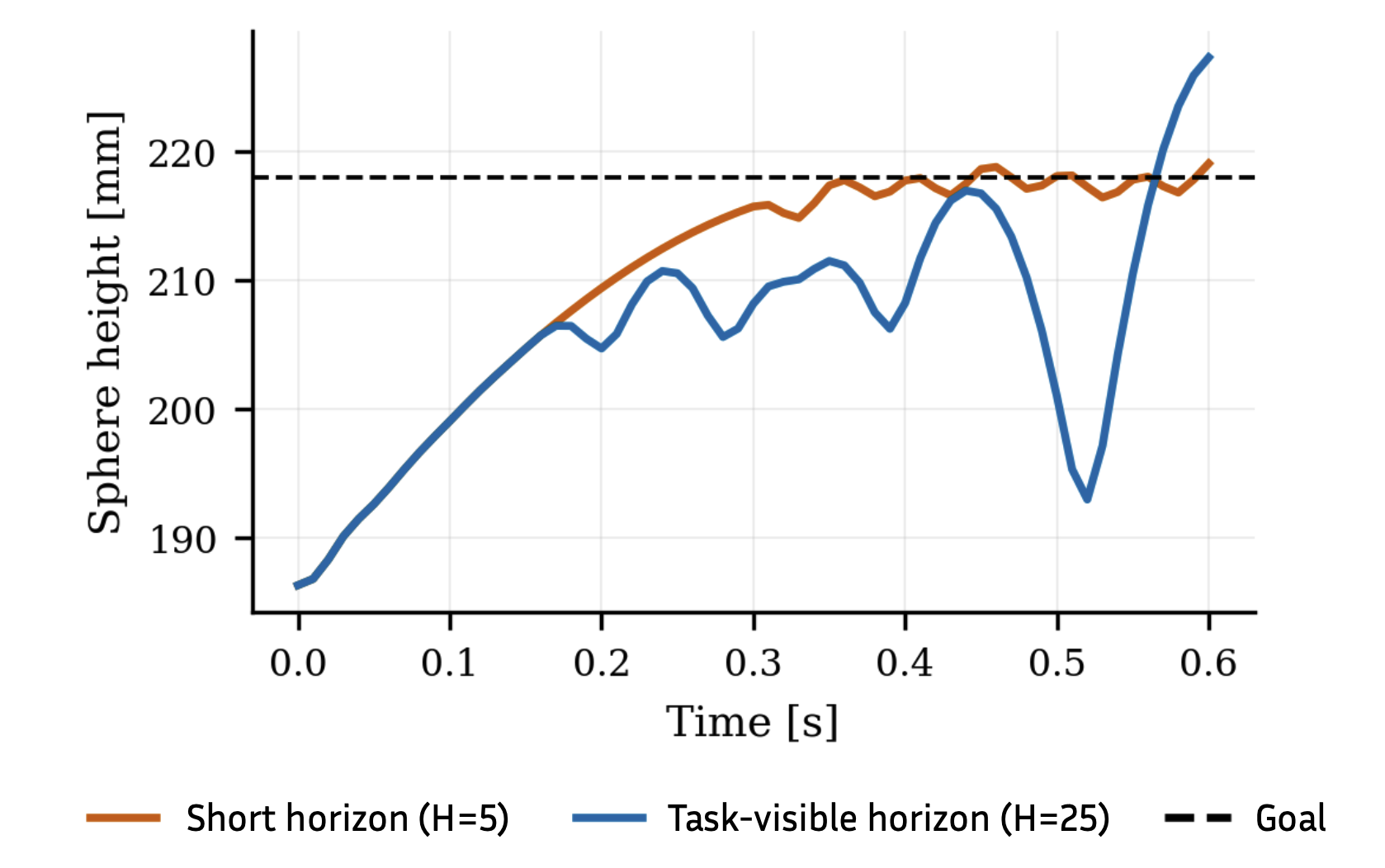}
    \caption{\textbf{Effect of the MPC planning horizon on the Tripod-Sphere task}. The method, number of rollouts per update ($N=8$, $R=3$), action parameterization, and execution conditions are identical. Only the horizon changes. The short horizon produces a quasi-static lift, whereas the task-visible horizon reveals the lowering-and-release strategy required for a jump.}
    \label{fig:tripod_MPC_horizon}
\end{figure}

\begin{table*}[!t]
  \caption{CPU and GPU rollout throughput across the benchmark. All measurements use $H=16$ and a parallel batch size of $64$, and exclude JIT compilation.}
  \label{tab:timing}
  \centering
  \begin{adjustbox}{max width=\textwidth}
    \begin{tabular}{r|ccccc}
      \toprule
      \textbf{Task} & \textbf{CPU batch latency} [s] & \textbf{GPU batch latency} [s] & \textbf{CPU} [rollouts/s]& \textbf{GPU} [rollouts/s] & \textbf{GPU/CPU speedup}\\
      \midrule
      \rowcolor{pastelblue}
      Cartpole & 0.282 & 0.095 & 226.7 & 673.3 & 2.97 \\
      Cart-Stem& 1.726 & 0.438 &  37.1 & 146.1 & 3.94 \\
      \rowcolor{pastelblue}
      Dynamic Trunk &  1.078 & 0.297 &  59.4 & 215.3 & 3.63 \\
      Tripod-Sphere & 7.311 & 3.140 &  8.8 & 20.4 & 2.33 \\
      \rowcolor{pastelblue}
      Three-axis Trunk & 18.754 & 3.372 &  3.4 & 19.0 & 5.56 \\
      \bottomrule
    \end{tabular}
  \end{adjustbox}
\end{table*}

The tripod experiment isolates the role of the planning horizon while keeping the method, rollout budget per update, action parameterization, and execution conditions fixed. Results are shown in~\cref{fig:tripod_MPC_horizon}. With $H=5$, the predicted trajectories do not extend far enough to expose the benefit of first lowering and subsequently launching the sphere. The controller therefore maintains the sphere near the platform and reaches a maximum sphere-platform gap of only $1.94$~mm. With $H=25$, after an initial direct motion of the platform, the complete interaction becomes visible within the planning window. The resulting controller then accepts a temporary decrease in sphere height and produces a launch with a maximum gap of $28.57$~mm. Replanning therefore does not remove the need for a task-visible horizon. Feedback corrects deviations from the current plan, whereas the horizon determines which delayed consequences can influence the action selected at the current MPC update.

\subsection{Computational performance}
\label{subsec:computation}

We evaluate the computational cost of the sampling process and the extent to which batched rollout evaluation can benefit from parallel execution. 

Timing measurements exclude the initial JIT compilation and are collected after a warm-up call. Each reported latency is averaged over five repeated executions of the same compiled computation. GPU timings include explicit device synchronization.
CPU timings are measured on an Apple M3 Pro with 36 GB of unified memory under macOS, and GPU timings on an NVIDIA A100-SXM4 with 80~GB of memory under Linux. The dynamic trunk is used to analyze batching. 

\begin{figure}[h]
    \centering
    \includegraphics[width=0.49\textwidth]{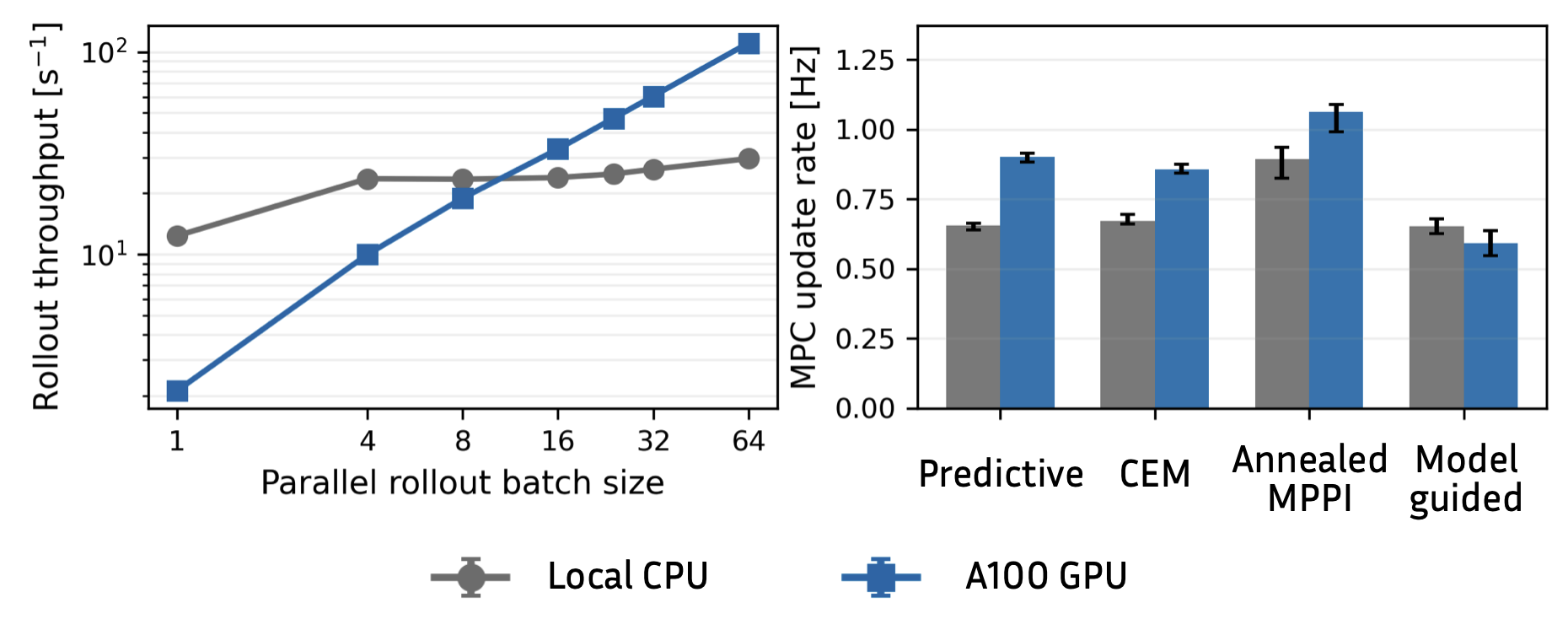}
    \caption{\textbf{Computational performance on the dynamic trunk}. Left: throughput in complete rollouts as a function of the number of action sequences evaluated in parallel. Right: complete MPC-update rate for $N=8$ samples and $R=3$ refinements. An MPC update includes sampling, rollout simulation, and distribution updates. Model-Guided MPPI additionally includes the local-model computations. In both cases, $H=25$.}
    \label{fig:computational_performance}
\end{figure}

The left panel of~\cref{fig:computational_performance} shows that the GPU provides lower throughput than the CPU for individual $25$-step rollouts because kernel-launch and dispatch overhead dominate the simulation time. Throughput then grows rapidly with the batch size, and the GPU overtakes the CPU around a batch of $16$ trajectories. At a batch size of $64$, the GPU evaluates approximately $215$ complete rollouts per second, corresponding to a $3.63\times$ speedup over the CPU. This scaling confirms that the main source of accelerator efficiency is parallelism across independent candidate trajectories. 

The right panel reports complete MPC-update rates for $H=25$, $N=8$ and $R=3$. At this smaller effective batch, the GPU remains partially underutilized and provides only a moderate acceleration for Predictive Sampling, CEM, and Annealed MPPI. Model-Guided MPPI is slower on the GPU than on the CPU in this configuration because its adjoint and local-model computations contain more sequential operations and expose less trajectory-level parallelism than the rollout batch. The computational benefit of accelerator execution therefore depends on both the number of parallel trajectories and the fraction of each update spent in the batched forward simulator.

The current simulator was not specifically designed for GPU execution. Several operations inside the implicit solve remain sequential, and the model-guided derivative computations do not yet exploit the same level of batching as the forward rollouts. Nevertheless, at a batch size of $64$ and a horizon $H=16$, GPU execution improves rollout throughput for all five environments, with speedups ranging from $2.33\times$ to $5.56\times$, as reported in~\cref{tab:timing}. The largest gain is observed for the Three-axis Trunk, whose more expensive trajectories better amortize device overhead. These results indicate remaining performance headroom from additional parallelization of the implicit solve and the model-guided derivative computations.

\section{Conclusion}
\label{sec:conclusion}

In this paper, we presented a comparative analysis of sampling-based control for deformable robots under limited rollout budgets. The study combines a benchmark of five representative environments with a common presentation of sampling strategies ranging from direct trajectory sampling to distribution adaptation, cost-weighted updates, and locally model-informed proposals. The benchmark spans stabilization, contact-mediated reaching, object interaction, elastic energy exploitation, and high-dimensional manipulation, and evaluates the methods in both open-loop and receding-horizon settings. Because trajectory evaluation relies only on forward rollouts, the comparative framework remains agnostic to the underlying deformable simulator and can be combined with any dynamics model that provides sufficiently efficient forward simulation.

Across all benchmark tasks, the experiments indicate that sample efficiency does not primarily arise from a universally superior sampling strategy. Instead, it results from introducing appropriate structure and information into the search. Low-dimensional temporal parameterizations, coarse-to-fine exploration, preservation of coherent contact modes, and local proposal guidance consistently improve performance under limited rollout budgets. The local model is most effective once sampling has already identified a promising mode, while the nonlinear simulator remains the reference for evaluating candidate trajectories. These mechanisms enable the controller to exploit contact transitions, delayed mechanical effects, and elastic energy within the limited rollout budgets imposed by deformable simulation.

The current study is limited to simulated environments and a penalty-based contact model. Future work includes learned proposal distributions, reduced-order and surrogate models for faster rollouts, adaptive planning horizons, improved contact formulations, greater exploitation of GPU architectures, and tighter integration with differentiable simulation. The benchmark and implementation presented here provide a reproducible basis for the systematic development and evaluation of future sampling-based control methods for deformable robots.

\section*{ACKNOWLEDGMENT}
This work was supported in part by the French government, under the management of the \textit{Agence Nationale de la Recherche} (ANR), as part of the “France 2030” program (grant ANR-23-IACL-0008, PR[AI]RIE-PSAI project; and grant ANR-22-EXOD-0008, PEPR O2R–AS4 project). It also received support from the European Union through the ARTIFACT project (Grant Agreement No. 101165695).

\balance{}
\bibliography{bibliography}
\bibliographystyle{ieeetr}

\end{document}